\documentclass[conference,letterpaper]{IEEEtran}
\pdfoutput=1
\usepackage{amssymb,amsmath}
\usepackage{setspace}
\usepackage{graphicx}
\usepackage{epsfig,cite,latexsym,subfigure}
\usepackage{theorem}
\usepackage{times}
\usepackage{epsf,cite,subfigure}
\usepackage{amsmath,amssymb}
\usepackage{floatflt}
\usepackage{url}
\usepackage{times}
\usepackage[usenames]{color}
\usepackage{dsfont}

\newcommand{\eqnref}[1]{(\ref{eqn:#1})}

\newcommand{\eqnlabel}[1]{\label{eqn:#1}}

\newcommand     {\paren}[1]{\left(#1\right)}

\newcommand{\norm}[1]{\left\|#1\right\|}
\newcommand{\curlb}[1]{\left\{#1\right\}}
\newcommand{\squareb}[1]{\left[#1\right]}

\begin{document}
\title{\Large Deep-Learned Compression for Radio-Frequency Signal Classification}

\author{
\IEEEauthorblockN{\normalsize Armani Rodriguez, Yagna Kaasaragadda, Silvija Kokalj-Filipovic$^*$}
\IEEEauthorblockA{\small Rowan University\\
\small\em rodrig52@students.rowan.edu, kaasar57@students.rowan.edu, kokaljfilipovic@rowan.edu$^*$}}

\maketitle
\begin{abstract}
Next-generation cellular concepts rely
on the processing of large quantities of radio-frequency (RF)
samples. This includes Radio Access Networks (RAN) connecting
the cellular front-end based on software defined radios (SDRs)
and a framework for the AI processing of spectrum-related
data. The RF data collected by the dense RAN radio units and
spectrum sensors may need to be jointly processed for intelligent
decision making. Moving large amounts of data to AI agents may
result in significant bandwidth and latency costs. We
propose a deep learned compression (DLC) model, {\em HQARF}, based on learned vector
quantization (VQ), to compress the complex-valued samples of
RF signals comprised of 6 modulation classes. We are assessing
the effects of HQARF on the performance of an AI model trained
to infer the modulation class of the RF signal. Compression
of narrow-band RF samples for the training and off-the-site
inference will allow for an efficient use of the bandwidth and
storage for non-real-time analytics, and for a decreased delay
in real-time applications. While exploring the effectiveness of the
HQARF signal reconstructions in modulation classification
tasks, we highlight the DLC optimization space and some open
problems related to the training of the VQ embedded in HQARF.
\end{abstract}
\vspace{-2mm}
\section{Introduction}\label{sec:intro} 
Data reconstruction from lossy compression incurs a loss of
information when the information rate in bits drops below the
theoretical lossless minimum, equivalent to the data entropy
\cite{shannon}. If a trained model is used to infer data properties from
the reconstructions that suffered information loss relative to
the training data, its performance may deteriorate \cite{Codevilla2021LearnedIC}, \cite{wiseML23}.
This paper considers digitally-modulated radio-signal samples
in the baseband, intended for the use by a remote deep learning
(DL) model trained to infer the signal modulation from such
samples. Next-generation (NextG) cellular concepts will rely
on the processing of large quantities of RF samples. This
includes the new Radio Access Networks (RAN) integrating 
the cellular front-end with 
the multi-access edge computing (MEC) architecture and
the RAN Intelligent Controller (RIC) framework for AI/ML
processing of the spectrum-related data. The RF data collected
by the RAN radio units of multiple adjacent NextG cells and spectrum sensors may need to be jointly processed
for intelligent decision-making. This may happen both at the
edge and in the cloud. Important architectural questions are
yet to be answered, including how the AI agents can access
the data and analytics from the RAN while minimizing the
overhead of moving them from the RAN to the storage and
inference locations. Moving large amounts of data results
in significant bandwidth and latency costs \cite{DataORAN}. We believe
that it is important to explore the possibility of RF data
compression that would preserve the utility of the data. We
here apply DL compression (DLC, or learned compression - LC) to compress
the complex-valued samples of RF signals comprised of 6 modulations classes. We are assessing the effects of such compression on the performance of an AI model trained to infer the modulation class of captured RF signals and then make intelligent decisions based on their classification. 

Machine-learning-based modulation classification is an important part of the RF machine learning (RFML) \cite{RFMLSurvey} used in spectrum management, electronic warfare, interference detection and threat analysis. Compression of baseband RF samples for the RFML training and off-site inference will allow for an efficient use of the bandwidth and storage for non-real-time analytics, and for a decreased delay in real-time applications. While exploring the feasibility of such a compression for the modulation classification task, this paper also highlights some open problems related to vector-quantization methods embedded in the LC training.
In this setup, an RF datapoint, which is an array of complex-valued narrowband RF samples, is to be reconstructed  from its compressed representation by the user of the classification model. The compressed representation will be received over a network or retrieved from a storage with no errors. We will refer to the inference task as remote, even though it may be local to where the data is generated and compressed, such as when the compression is motivated by the limited storage space.  
It is convenient to use the 'remote' identifier to avoid confusing it with the RFML model for {\em learned compression} (LC) that we propose here. 

{\bf Prior and Proposed LC work:} An LC model is trained to seamlessly compress data using DL algorithms. LC may leverage discriminative models such as autoencoders \cite{scalableAE}, or generative models such as variational autoencoders (VAE) \cite{Kingma2013AutoEncodingVB} and generative adversarial networks (GAN) \cite{Mentzer2020HighFidelityGI}. The most popular LC architectures typically include a neural net backbone built upon the VAE architecture \cite{LCreview}. One of the latest deep compression models, known as VQ-VAE \cite{VQVAEnips}, is an extension to VAE that employs learned vector quantization (VQ). For 30 years, since \cite{LVQpack}, the learning of optimal VQ codebooks has been an open problem resulting in many attempts to generate a converging algorithm that could learn the quantization vectors for any type of data. The LC proposed here, {\em HQARF}, will be   analyzed using a family of models, starting from a hierarchical autoencoder, trained using only the reconstruction loss, via an extended model that performs vector quantization (VQ) of the autoencoder's latent space, and ending with a generative model, like VQ-VAE, whose generative loss compares the posterior of the quantized latent representation with a categorical prior. The generative model is motivated by the possibility to leverage statistical diversity of reconstructions to mitigate reconstruction loss and adversarial attacks \cite{goodfellow2015explaining}, \cite{szegedy2014intriguing}, \cite{gu2015deep}. The trainable VQ codebook \cite{Quant,LVQpack} helps to achieve a desired compression rate while maximizing the task-based  utility of the reconstructions. To allow for scalability, HQARF maintains a hierarchical architecture. This hierarchical architecture is based on \cite{hqa}  in which a hierarchical version of VQ-VAE, called Hierarchical Quantized Autoencoder (HQA) has been applied to simple image datasets. 
To the best of our knowledge, learned compression has not yet been applied to the RF data. We will therefore first explore it in a small, task-specific context, aiming to assess how lossily compressed RF data affects the  accuracy of an R2FML modulation classification model whose training dataset did not include lossy compression. We motivate the problem and define the basic model in Sec.~\ref{sec:sys}. We discuss the details of the compression architecture and the training process, including the loss functions in Sec.~\ref{sec:comp}. The classification model and the evaluation of the HQARF reconstructions are discussed  in Sec.~\ref{sec:class}. We conclude in Sec.~\ref{sec:concl}.  
\vspace{-2.5mm}
\section{System Model}\label{sec:sys} 
\begin{figure*}
  \includegraphics[width=16.0cm,height=3.8cm]{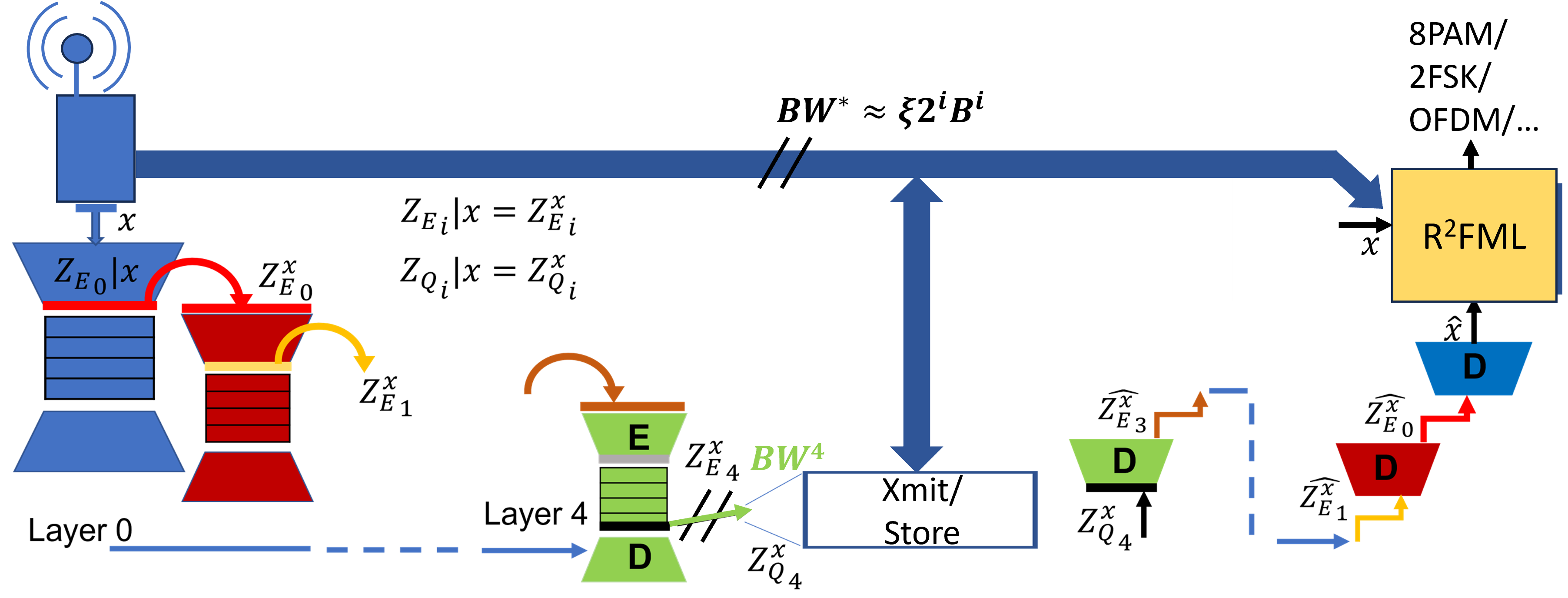}
	\vspace{-4mm}
  \caption{Shown is the information flow from a SDR through HQARF compression to layer 4 (L4). L4 requires the bandwidth  $B^4,$ to store or transmit the compressed information about the datapoint $x$ composed of 1024 complex-valued samples, as opposed to directly storing/transmitting $x$ for a remote classifier to infer its modulation class. If a compressed representation $Z_{Q_i}, i \in \curlb{0,\cdots,4}$ is stored/ transmitted, the same HQARF  model is used to recover $x,$ decompressing $Z_{Q_i}$ into $\hat{x}.$ We measure the effectiveness of compression by comparing the accuracy of R2FML between the reconstruction $\hat{x}$ and the original $x$ for various hierarchical compression rates;currently, $\xi= 1.37$.} \vspace{-4mm}
\label{fig:scen}
\end{figure*}
The hierarchical nature of HQARF allows us to use the same compression model adaptively for different compression rates, and analyze the effectiveness of the quantization on different levels. Multiple compression rates may be important for joint network source coding.  Fig.\ref{fig:scen} depicts our system model where after the compression is done by HQARF, the compressed representation from the desired level (or multiple levels) is sent to a R2FML classifier (or stored, awaiting  retrieval by the classifier). The reconstruction is performed at the remote site using the same trained HQARF to recover the original data before classifying it by the R2FML model. The reconstruction uses as many hierarchy layers as the compression has used.  Our HQARF model made 2 significant modifications to HQA. {\bf First},  we modified the architecture to work with vectors of complex-valued RF samples instead of images, and modified the reconstruction loss to account for the complex phase reconstruction. {\bf Secondly}, we took a hierarchical  approach to training and analyzing the model; first, we train a hierarchical autoencoder (HAE), then we transfer-learn a vector-quantized version of that model using the trained weights of the HAE, while adding a trainable VQ codebook to quantize the HAE bottleneck, accompanied by a loss component that measures the quantization error; finally, we add the generative loss component based on the Kullback-Liebler divergence, effectively creating a hierarchical VQ-VAE for the RF data (HQARF). 
\vspace{-2.5mm}
\section{Hierarchical Vector-Quantized Compression of RF Datapoints}\label{sec:comp} 
\vspace{-2mm}
{\it Lossless compression} is about finding the shortest digital representation (in bits) of a signal. Lossless compression algorithms take as an input arbitrary information signal represented (sampled) as a sequence of $N$ symbols and process it with the objective to find its shortest compressed representation: a sequence of uniformly distributed bits which cannot be compressed further without a loss of information. Consequently, lossless compressed representation allows for complete reconstruction of the original sequence of $N$ symbols. Information theory  sets the foundations for entropy coding, with the length $n$ of the shortest lossless representation equal to the signal’s entropy, resulting in a rate $R=n/N$ bits-per-symbol. {\it Lossy compression} achieves an even shorter representation and lower rate $R$ but the signal reconstructed from such representation suffers a distortion $D$ from the original signal. However, certain utilities of the signal reconstructed from the lossy compression may be unaffected. For example, lossy reconstructed data with a distortion $D_u$ may still be fully classifiable by a deep learning model that was trained on uncompressed data.

Using HQARF to generate lossy reconstructions, we will analyze the effect on the classification accuracy depending on the compression rate $r_\bullet$ (the size in bits relative to the original size). We will compare these with the original of the unit compression ratio. Here, different compression rates $r_i$ are expected to match different bandwidths under a low-latency transmission $\tau_{RF}$, and/~or different storage capacity. Please see Fig~\ref{fig:scen} where the original $x$ requires bandwidth $\geq B_{\ast}$ to be transmitted to the remote classifier within latency $\tau_{RF}$ while the HQARF hierarchy levels compress $x$ to fit the bandwidth $B_{i} < B_{\ast},$  where $i\in \curlb{0,\cdots 4}$ is the hierarchy index. To be able to analyze the feasibility of a given classification accuracy under the constraint $B_i$, let us explain the compression model and the methodology of its training in detail.
\vspace{-2mm}
\subsection{HQARF Dataset for modulation recognition (ModRec)}\label{subsec:lossy}
\vspace{-1mm}
Consider the problem of inferring a property of a physical signal from the signal’s reconstruction 
$\hat{x}$ out of a lossily-compressed representation.  We narrow down that question to the classification accuracy in the deep learning setup, given the compression rate. We denote by  $MR_{\theta}(x)$ the R2FML algorithm for modulation classification (popularly known as modulation recognition - {\em ModRec}), where the weights $\theta$ have been trained on $\curlb{x \in X}.$
We are interested in comparing $A(\hat{x})$ and $A(x)$, where $A(\bullet)$ is the accuracy of $MR_\theta.$  The datapoint $x$ can be represented as $x=\squareb{Re_i+j Im_i}, i = 1\cdots p$ with $j=\sqrt{-1}$.
Let us describe how $x$ is created from a modulated signal
$u,$ obtained as $u = M_s(b)$, where $s \in \mathcal{S}$ is the employed
modulation scheme, and $b$ are information bits. $S$ denotes the finite set of available
digital modulation schemes. In our experiments, $$\mathcal{S} = \squareb{4ask,8pam,16psk,32qam-cross,2fsk,ofdm256},$$ so we refer to our dataset as {\em 6Mod}. $M_s = \curlb{0,1}_m \rightarrow \mathcal{C}_n$ describes the modulation function. The sequence of bits $b=\curlb{0,1}_m$ is encoded into a sequence of complex valued numbers of length $n,$ where the complex sample $c_i=Re_i +j Im_i,$ encodes the modulation phase $\phi=\arctan{Re_i/Im_i},$ and amplitude $a_i = \sqrt{Re_i^2+Im_i^2}.$ We create datapoints as sub-sequences  $x$ of $u \in \mathcal{C}_n,$ of length $p = 1024,$ which depending on the modulation contain more or less mappings of the original random sequence of bits $b.$ This leads to an imbalanced approach to classifying modulations because in any given sequence of length $p$ we will see more randomness due to the original random bit sequence $b$ in low-order modulations than in high-order modulations. However, as this is the common approach in the RFML ModRec, we do not consider its effects on the classification accuracy even though the selected $\mathcal{S}$ contains diverse modulation orders (how many original bits are represented by a single complex value). 

We prepared a synthetic modulation dataset by using the open-source library {\em torchsig} featured in \cite{torchsig}. The torchsig library here emulates the clear-channel samples of high SNR while the effect of the channel and receiver imperfections will be addressed in future research. The library function {\em ComplexTo2D} is used to transform vectors of complex-valued numbers into the the 2-channel datapoints, with each channel comprised of $p$ real numbers, previously normalized. Channel 1 contains real components {\em(i)} and channel 2 the imaginary ones {\em(q)}. The fact that our datapoints are 2-D vectors of real numbers required modifications of the architecture in \cite{hqa} (see Section~\ref{subsec:nn}).
\vspace{-3mm}
\subsection{Generative Deep Learned Compression of RD datapoints using hierarchical VQ-VAE}
\begin{figure}[h]
\vspace{-1mm}
\centering
\hspace{-1mm}\includegraphics[width=0.41\textwidth]{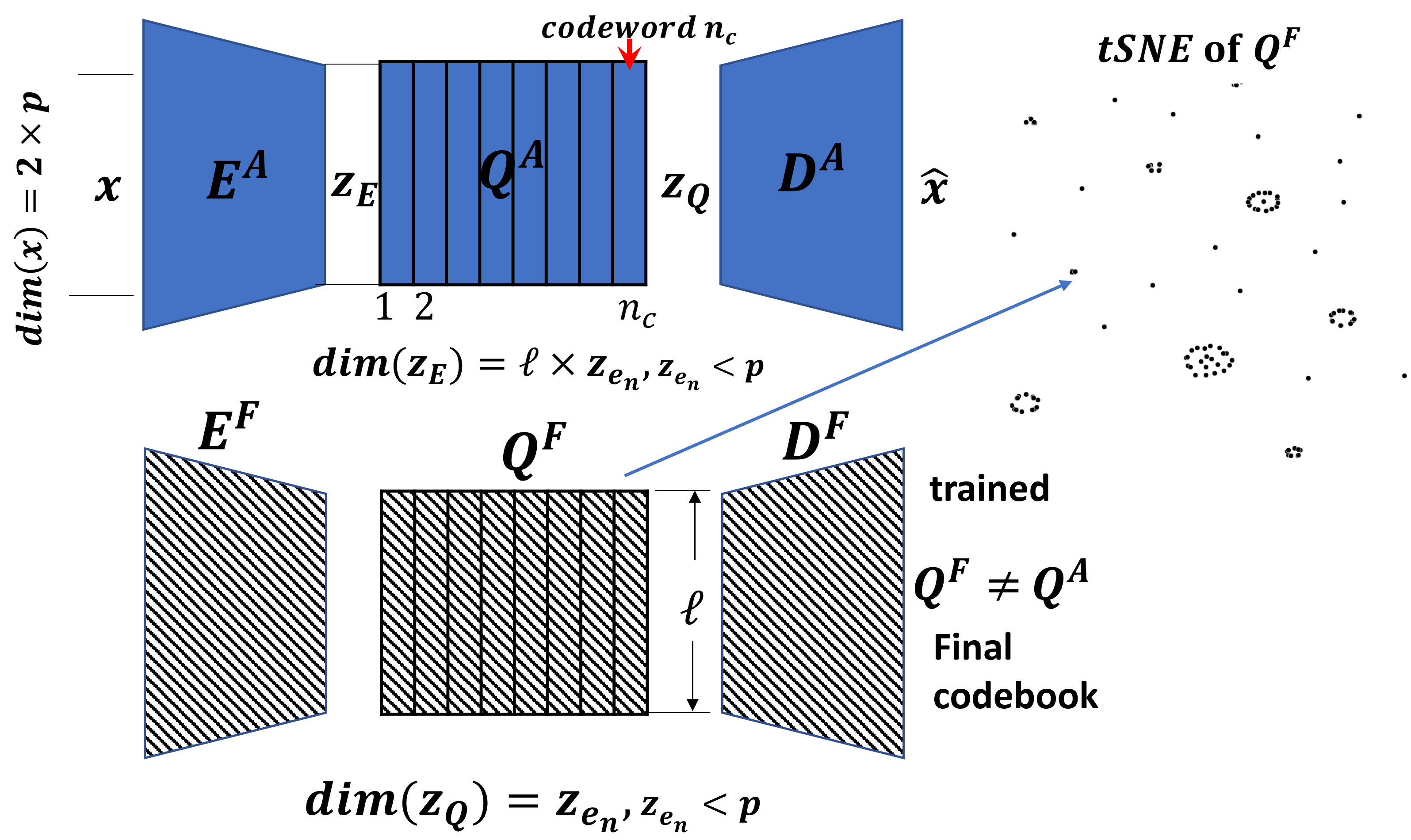}
\vspace{-3mm}
\caption{The training of VQ-VAE - {\bf top:} randomly initialized parameters of the encoder (E), decoder (D) and the $n_c$ quantization codebook (Q) vectors of dimension $\ell$;  {\bf bottom:} the final trained VQ-VAE where a single codeword's index from the trained Q will be associated with each of the $z_{e_n}=dim(z_e)[1]$ slices of $x$'s  latent projection $z_e$. Therefore $x$ is compressed to $z_{e_n} \times \log_2(n_c)$ bits. {\bf On the right:} The t-SNE visualization of the trained Q shows clusterization around a few codewords. 
}\vspace{-3.5mm}
\label{fig:vqvae}
\end{figure}

\vspace{-1mm}
The HQARF uses a hierarchy of VQ-VAEs in which the encoder's output of the first layer (L0) $z_e$ (creating the least compressed reconstruction) is the input into the second VQ-VAE and so on (Fig.~\ref{fig:scen}). The layers are numbered 0 to 4 where the $i$th $z_e$ is of dimension $dim(z^i_e) = (\ell,p/2^{i+1})$. The Vector-Quantized Variational Autoencoder (VQ-VAE) model is a generative unsupervised machine learning algorithm that builds upon the VAE model \cite{KingmaVAE} through the use of a vector-quantized, discrete latent space $z_q$ as illustrated in Fig.~\ref{fig:vqvae}. Architecturally, a VQ-VAE is composed of 3 modules: {\bf E} - the Encoder neural net (with output $z_e$),  {\bf Q} - the Vector-Quantizer (with output $z_q$) and {\bf D} - the Decoder net which produces the reconstruction of the input $x$, denoted $\hat{x}$. The latent representation $z_q$ lends itself well to producing higher-quality input reconstructions relative to standard VAE models (composed of E and D only), as proved in the computer vision domain. Moreover, $z_q$ produces lower information rate $I_{z_q}$. 
The hierarchy of the Encoder-Decoder (E-D) blocks (representing an autoencoder), which is of the same architecture as the respective HQARF blocks, but trained without the Q block and a generative loss, is denoted here as HAE. Let us refer to the outermost level of HQARF as VQAE0 and the same architecture without the  {\bf Q} block as AE0. The output $z_e$ of the $E$ block, is of the same dimension in VQAE0 and AE0, which does not have to be lower than the input's dimension. The compression is achieved  by adding the $Q$ block which projects $z_e$ into $z_q.$ In fact, the $z_e$ in the VQAE0 (AE0) of the HQARF showcased here projects the input $x$ of dimensions $2\times 1024$ into $z_e$ of dimensions $\ell \times z_{e_n}, $ where $dim(z_e)[0] = \ell=64,$ and $dim(z_e)[1] = z_{e_n} = 512.$ Obviously, AE0 acts more like a hyperdimensional vector encoder \cite{aygun2023learning} than a compression model.  
Due to the complexity of training all 3 components (E+D+Q) simultaneously, we performed the following ablation study. We first train the HAE, using the reconstruction loss $L_R(x, \hat{x}),$ and then transfer its learned weights to the respective blocks of the HQARF. Next, we train HQARF using a modified loss including the additional component which measures the quantization error, the commitment loss $L_Q = E_{q(z_q=k|x)}{\norm{z_e(x) - e_k}^2},$ where $e_k$ is the codeword $k$ of the quantization codebook (please see \eqnref{post} for the definition of the posterior $q(z_q=k|x)$). Note that every hierarchy layer trains a separate E, Q and D block.
Finally, after training this hierarchical vector-quantized HAE, we add a generative loss function and retrain HQARF to its final version. The generative loss is a Kullback-Liebler divergence between the posterior $q(z_q=k|x)$ and the categorical prior with $n_c$ classes, where $n_c$ is the number of codewords in Q.  

Information rate of the compressed representation $z_q$ can be calculated as $I_{z_q}=z_{e_n} \times \log_2(n_c),$where $z_{e_n}=dim(z_e)[1].$  The $n_c$ codewords (vectors) $e_j, j\in {1,\cdots,n_c}$ in the quantization codebook $Q$ are of dimension $\ell.$  Hence, for VQAE0's $z^0_e$, each one of its $z_{e_n}=512$ slices of dimension $\ell=64$  will be represented by a number, indexing a single codeword $e_j$ out of the $n_c=64$ codewords. Thus, for each of 512 slices the reconstructing user receives an index, losslessly represented by $\log_2(n_c)$  bits.  This is possible as the Q  consists of $n_c=2^6=64$  codewords of of dimension $\ell = 64,$ while we are parameterizing the E architecture by the tuple $(\ell,h),$ to yield the dimensionality of the latent slice $dim(z_e)[0]=\ell.$  We explain the effect of $h$ in Sect.~\ref{subsec:nn}. 

The $z_q$ in other layers of HQARF will be quantized similarly as we make sure by the architecture design that each layer's $dim(z_e)[0]=\ell,$ equal to the codeword length. Note that codebooks across layers are of the same size $64 \times 64$ (although they do not have to be, as these are completely independently trained codebooks). Also, as far as their training is concerned, the codebooks are agnostic of  where the training data is coming from.  The optimal Q dimensions are one of the open questions that will be explored in future research. 
As our datapoints consist of 2 channels of normalized real and imaginary components \cite{torchsig}, we count $2p$ data elements, and for simplicity we consider each element to be independently drawn from a normal Gaussian distribution. It is known that Gaussian has the largest entropy $H_N(X)$ of all distributions of equal variance. For unit variance, $H_N(X)=1/2\log(2 \pi e)=2.05$. Hence, the information, expressed in the number of bits for each input $x$ is $I_x=dim(x)H_N(x)$. Recall that $z_q$ is  the quantized version of the bottleneck $z_e,$ and it follows multivariate categorical distribution of size $n_c,$ as each $z_e[1]$ elements (slices of $z_e$) will be represented by one of the codewords $e_j$ in the Q of size $n_c.$ Hence, $z_q$'s dimension is just $dim(z_e)[1]=z_{e_n}$, and each of the $z_{e_n}$ elements is described by $\log_2(n_c)$ bits. For $n_c=2^d$, the {\em compression ratio} will be  
\begin{align}
CR &= \frac{I_x}{I_{z_q}}=\frac{dim(x)H_N(X)}{z_{e_n}\log_2(n_c)} = \frac{2.05dim(x)}{d\times z_{e_n}}. \eqnlabel{CR}
\end{align} 
Hence, if  $d$ is such that $d\leq  = H_N(X),$ $CR \geq 1$ for each $z_e$ with $dim(x) > z_{e_n}),$ meaning that HQARF's $z_q$ compresses $x.$ We want to allow for a larger codebook to be able to perform good vector quantization training:  if instead of $n_c = 2^{2.05} \approx 4,$ we use $n_c=2^6$ ($d=6$), we need to design the 2nd dimension of $z_e$ to be significantly lower than $dim(x)$ in order to achieve good enough compression. Under this premise, we design the architecture of the $E-D$ on each hierarchy level to give us $z_{e_n} = dim(input)[1]/2.$ Here, {\em input } is the input to that E-D level. Hence, for L0, we have $CR_0=\frac{2.05\times 2\times p}{6\times p/2}=1.37,$ since  $dim(input)=dim(x) = 2\times 1024,$ and $z_{e_{n(0)}} = 512.$
The codeword index per element of the $2$nd dimension of $z_e$  is all that we transmit (store) on any compression level, given the user's knowledge of the trained codebooks. 
For any other level $i>0,$  input has the same number of channels $dim(input)[0]=\ell$ as the bottleneck, hence,
\begin{align}
\nonumber CR_i&=\frac{2.05dim(x)[0]\times dim(x)[1] }{6dim(z_{e_{(i-1)}})[1]/2}\\
\nonumber &=\frac{2.05\times 2\times p}{6dim(z_{e_{(i-2)}})[1]/2^2}\\
 &=\frac{4.1\times p}{{6\times p}/2^{i+1}} =\xi\times 2^{i},
\end{align}
where $\xi = 1.37$ is featured in Fig.~\ref{fig:scen}.
The above calculations yield the {\em compression rate} $r_0=1/CR_0 =0.73$, and each subsequent reconstruction's dimension is decreased to $1/2$ of the previous, resulting in $r_4 =0.73/16\approx 0.045$.
\vspace{-2mm}
\subsection{Neural Net architecture of VQ-VAE in HQARF} \label{subsec:nn}
\vspace{-1mm}
The encoder architecture for $z_{e_n}  = p/2$ is composed of 3 1-D convolutional layers, and the decoder consists of an equal number of 1-D deconvolutions. Despite the simple architecture of the E-D, the hierarchy, the stochastic component in the loss and its diverse structure, and the data structure made the training  difficult before we introduced a novelty: the best results were achieved by first training a hierarchy of autoencoders, using a 2-component reconstruction loss, and then transfer-learning the hierarchy of VQ-VAEs by transferring the weights of the encoders and decoders. The bottleneck $z_e$ is the output of third 1-D convolutional layer in E, with $\ell$ output channels. The other convolutional layers have the number of output channels affected by the HQARF parameter $h$, and that is how the learning capacity (number of weights) is controlled across the layers. We started with the parameter values inherited from \cite{hqa} but, testing them on the HAE hierarchy, we realized that these parameters are not optimal. Our criterion for optimality is based on the comparison of the evaluated classification accuracy $A_i(\hat{x})$ of the $L_i$ reconstructions and the accuracy that we expect based on the singular value decomposition (SVD) that we performed on the original data.
\begin{figure}[h]
\vspace{-2mm}
\centering
\hspace{-1mm}\includegraphics[width=0.37\textwidth]{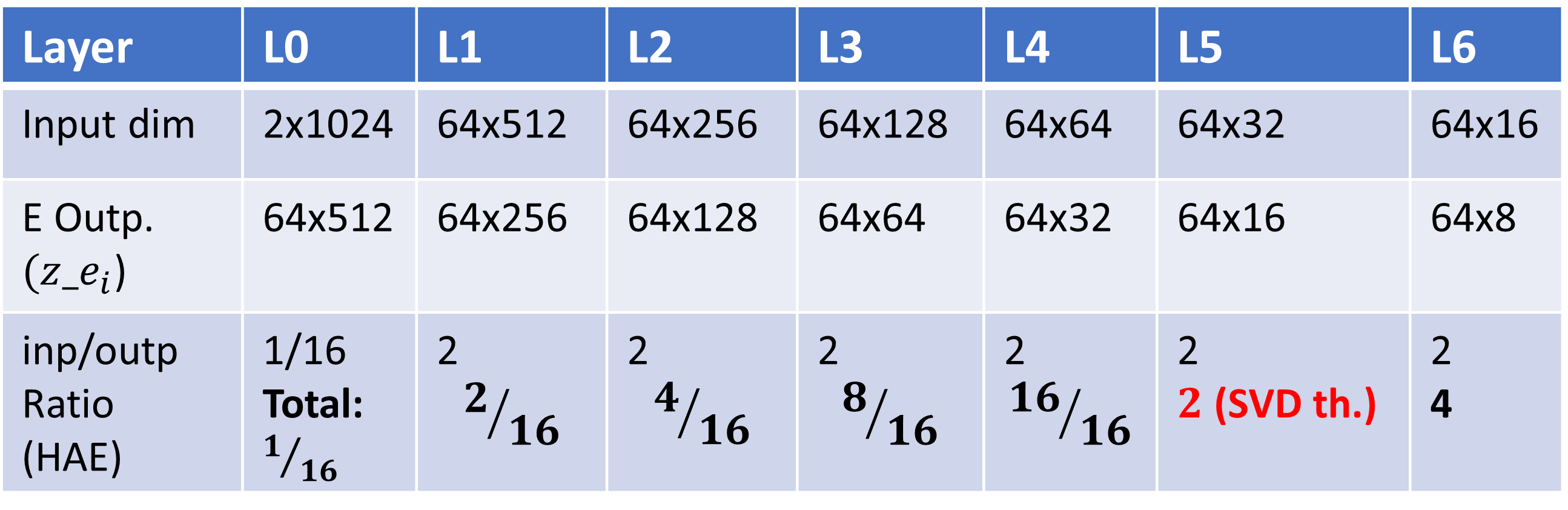}
\vspace{-3mm}
\caption{Table of the HAE encoders' input/ output dimensions and the SVD threshold (L5) for optimal $h$ parameters. 
}\vspace{-3mm}
\label{fig:tab}
\end{figure}

{\bf SVD-based threshold:} We performed an SVD on the original 6Mod data in the complex-valued domain, and calculated how many eigenvectors (significant dimensions) we should keep to preserve more than 99 \% of the total information in the data. The result is that we need 500 out of the original 1024 complex eigenvectors. This means that designing $z_e$ s.t. the product of its dimensions is equal to $0.5\paren{dim(x)[0]\times dim(x)[1]}$, allows for the $\hat{x}$ reconstructed from such a $z_e$ to be perfectly classified. Hence, according to the table in Fig.~\ref{fig:tab}, L5 is our achievability reference, as its $z_e$ has 1/2 of the original dimensions: if we manage to achieve the $A_5=100\%$ accuracy with L5, it means that the HAE E-D chain is parameterized well (and so is HQARF). With original parameters, $A_4$ was as low as 50\% while it is now $80$ (ref. Fig. \ref{fig:acc}). Based on the SVD threshold, we started modifying those hyperparameters and achieved the current performace (Fig. \ref{fig:acc}), meaning that we are close to the best performance but not yet there. 

The Q is designed as a learnable tensor of dimension $n_c \times \ell,$ s.t. we can train it based on the MSE distance $d_{MSE}$ between each codeword of length $\ell,$ and each of the $z_{e_n}$ slices of length $\ell$ (slice is the column partition of $z_e$).  As in \cite{hqa}, we pick the codeword to quantize each slice using a stochastic method, based on sampling the posterior probability 
\begin{align}
q(z_q = k|x) = \exp^{-\norm{z_e(x) - e_k}^2}. \eqnlabel{post}
\end{align}
The codewords (CWs) are being learned starting from random Gaussian samples at the initialization, and converging to a Q that minimizes the loss function, composed not only of the reconstruction loss $L_R(x, \hat{x}),$ but also a generative loss between the posterior and a categorical prior, and a commitment loss measuring the distance between the $z_e$ and the chosen $e_k.$ Note that, in the outermost layer L0, we added a new component to $L_R(x, \hat{x})=L_{MSE} + L_{\phi}$, to measure not only the MSE distance between $x$ and $\hat{x},$ but also the cosine loss $$L_{\phi}=1/p\sum_{i=1}^{p}{\frac{x[i,:]\times\hat{x}[i:0]^T}{\norm{x}\times\norm{\hat{x}}}}.$$ As $x[i,:]$ are the real and imaginary parts of the $i$th sample, $L_{\phi}$ measures the phase reconstruction, a very important feature in digital phase modulations.  For details of the Q training, please consult our code \cite{hqarfcode}. Apart from the typical tuning of the Q parameters using stochastic gradient descent of the loss, and obtaining a differentiable
sample from the posterior \eqnref{post} via the Gumbel Softmax relaxation, the least used CW is periodically reset to the vicinity of the most used CW. We considered the reset period to be a hyper-parameter and obtained good results when it increased, as frequent resetting foster instability. More importantly, we defined the vicinity of the CW adaptively, circling in with the number of resets (see t-SNE \cite{tsne} visual of the codebook in Fig.~\ref{fig:vqvae}). The optimal reset policy will be determined in future research. 
\begin{figure}[h]
\vspace{-4mm}
\centering
\hspace{-1mm}\includegraphics[width=0.44\textwidth,height=4.8cm]{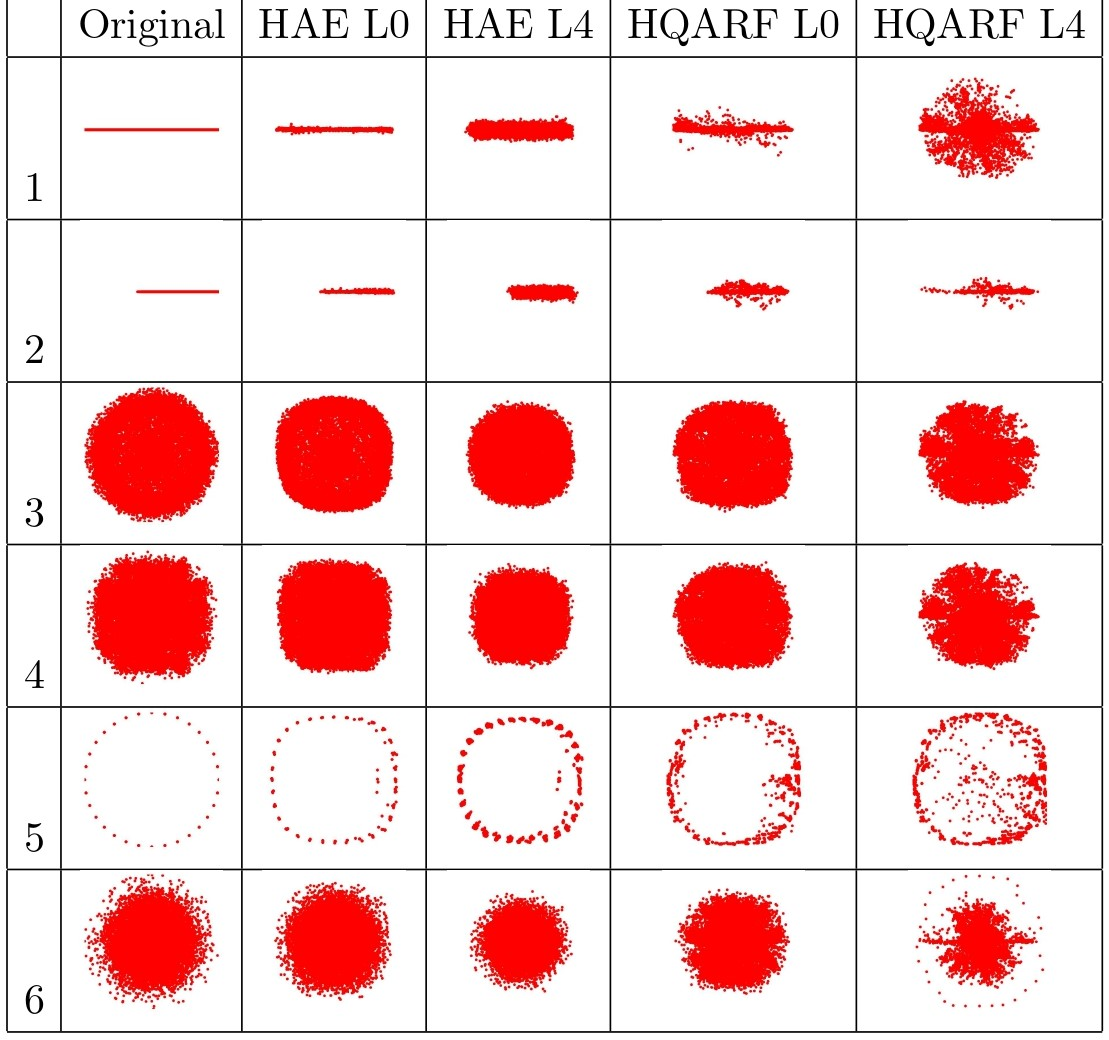}
\vspace{-3mm}
\caption{i/q scatterplot of 6 different classes based on the reconstructions across layers compared with the ideal (original) scatterplot. We concatenated 20 reconstructions of random datapoints of the same class, each comprised of 1024 complex-valued samples, and plotted them in the complex plane. 
}\vspace{-3mm}
\label{fig:constel}
\end{figure}
\begin{figure}[h]
\vspace{-0.5mm}
\centering
\hspace{-1mm}\includegraphics[width=0.46\textwidth,height=3.2cm]{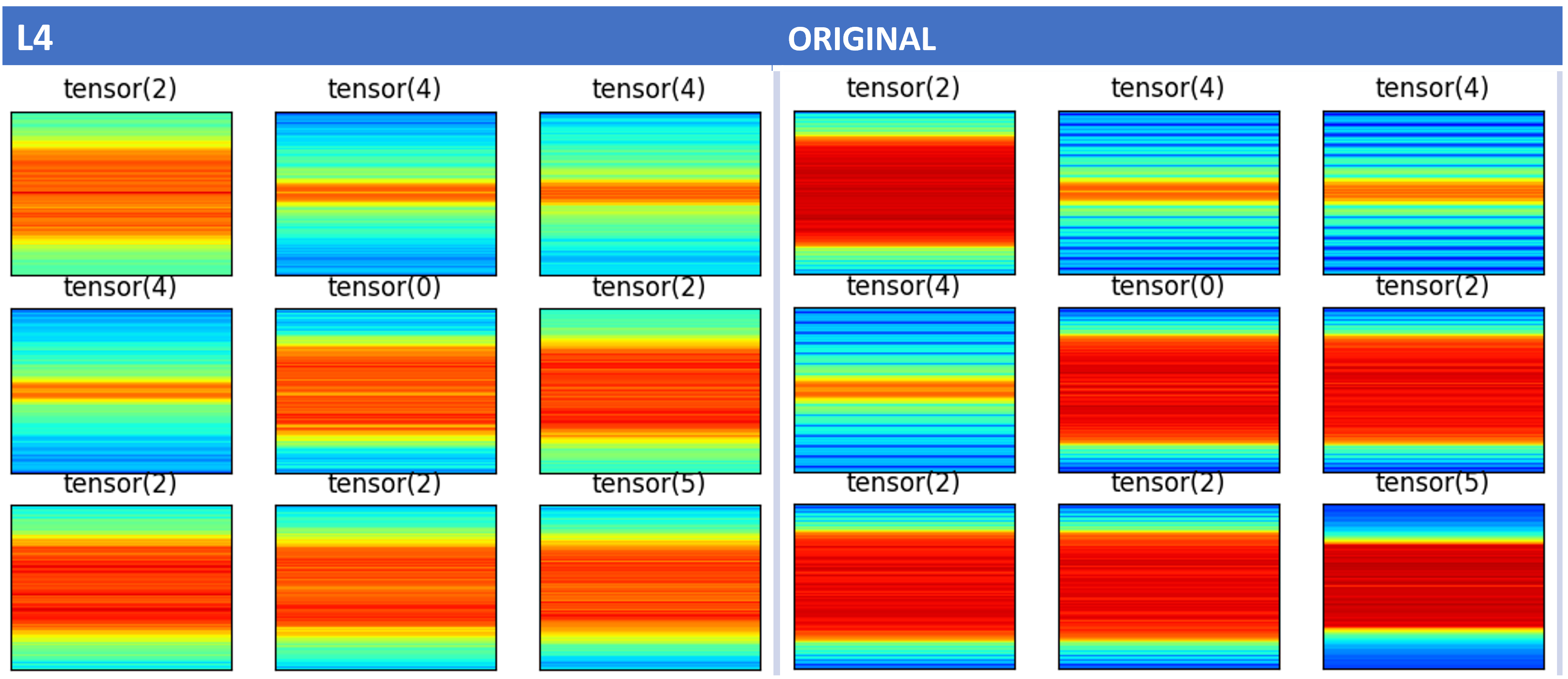}
\vspace{-3mm}
\caption{Spectrograms of L4 reconstructions and their originals for randomly selected modulations (0-5) 
}\vspace{-4mm}
\label{fig:spec}
\end{figure}
\vspace{-3mm}	
\section{Evaluation with the EfficientNet Classifier} \label{sec:class}
\begin{figure}[h]
\vspace{-4mm}
\centering
\hspace{-1mm}\includegraphics[width=0.44\textwidth,height=3.8cm]{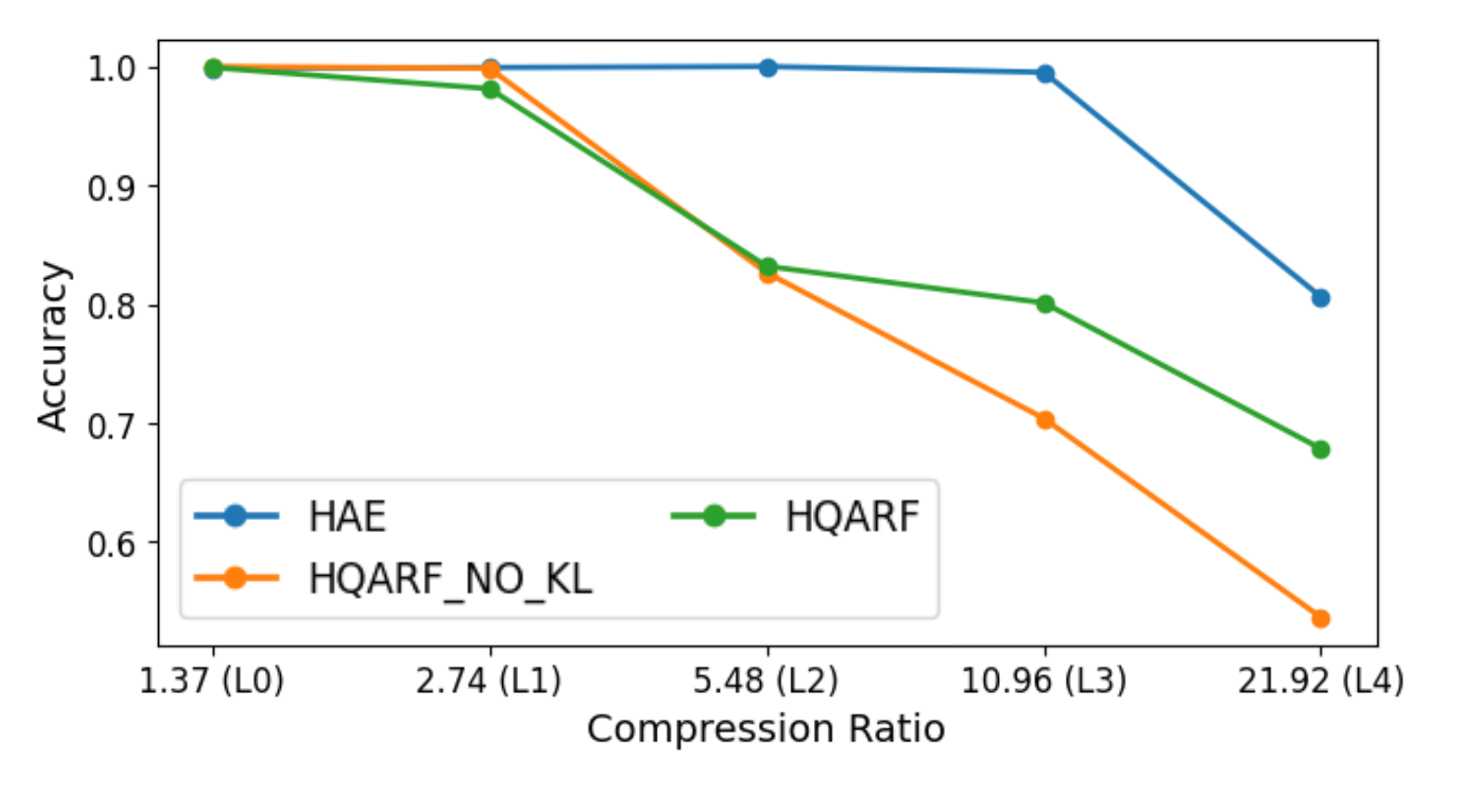}
\vspace{-6mm}
\caption{Accuracy vs compression ratio (CR) across Layers for HAE, HQARF\_NO\_KL and  HQARF  with Q of size $64 \times 64$.  The CR on the x axis {\bf does not apply to HAE}, as HAE does not perform VQ: HAE is added because by tracking how close we are to the  bound given by SVD, we know that we can do better than this result. 
}\vspace{-3.5mm}
\label{fig:acc}
\end{figure}

\vspace{-1mm}	
Upon training the 5 HQARF Layers on the {\em 6Mod dataset}, we evaluated it on the version of the {\bf EfficientNet\_B4} \cite{EffNet} used in \cite{torchsig}, which was appropriately transfer-learned, and evaluated on the original {\em 6Mod} dataset, resulting in a reference accuracy $A(x) \approx 100\%.$ Note that we trained and evaluated HAE with 7 layers (Fig.~\ref{fig:tab}) to be able to assess how well the architecture of the E+D is parametrized, while HQARF was trained for 5 layers.  Fig.~\ref{fig:acc} shows how the accuracy of reconstructions depends on the compression ratio (CR); here the HAE accuracy does not exhibit such CR but only illustrates that the space of the $h$ parameter, and possibly the architecture, should be further explored. Fig.~\ref{fig:constel} shows "`digital constellations"' of the originals and their reconstructions, and Fig.~\ref{fig:spec} shows random spectrograms of datapoints and their L4 reconstructions. While the real constellations show complex samples at symbol times, ours are the scatterplots of complex samples at a much higher rate obtained by baseband sampling.   However, they illustrate  gradual deterioration in the phase reconstruction, the same  as Fig.~\ref{fig:spec} in frequency domain, while the ModRec utility stays  good until L3/ L4 (Fig.~\ref{fig:acc}). 
\vspace{-2.5mm}	
\section{Conclusions and Future Work}\label{sec:concl} 
\vspace{-1mm}	
We introduce HQARF, the first vector-quantization (VQ) based learned compression (LC) of modulated RF signals and evaluate their lossy reconstructions on a modulation recognition (ModRec) task, illustrating  the utility of LC in this domain and its optimization space. The simple architecture and compact size of HQARF are very convenient for the quantization close to the radio interface. We point out to the complex factors affecting the ModRec accuracy on the HQARF reconstructions, and the fidelity of their complex-plane scatterplots and spectrograms. These factors include the HQARF architecture, training methodology, loss functions and the dimension and training of the VQ codebook. We defined a bound for the LC performance based on SVD. Pursuing this bound, we now have better results than presented in the paper. This is a proof of concept deserving further investigation, as it may have applications in intelligent network optimization and spectrum management. 
\vspace{-1mm}
\bibliographystyle{IEEEtran}
\bibliography{hqasigbib}
\end{document}